\documentclass[letterpaper, 10 pt, conference]{ieeeconf}  

\IEEEoverridecommandlockouts                              

\usepackage{graphics} 
\usepackage{epsfig} 
\usepackage{mathptmx} 
\usepackage{times} 
\usepackage{amsmath} 
\usepackage{amssymb}  

\usepackage{threeparttable}
\usepackage{textcomp}
\usepackage{siunitx}

\usepackage{adjustbox}
\usepackage{diagbox}
\usepackage{booktabs, multirow}
\usepackage[utf8]{inputenc}

\usepackage{adjustbox}
\usepackage{booktabs, multirow}
\usepackage{xcolor}
\usepackage[utf8]{inputenc}
\usepackage{graphicx}
\usepackage{tikz}
\usepackage{listings}                       
\usepackage{algorithm, algorithmic}                      
\usetikzlibrary{backgrounds}

\usepackage[noadjust]{cite}

\usepackage{gensymb}

\title{\LARGE \bf
OV-MAP : Open-Vocabulary Zero-Shot 3D Instance Segmentation Map for Robots}

\author{Juno Kim$^{1*}$ Yesol Park$^{1*}$ Hye-Jung Yoon$^{1*}$ Byoung-Tak Zhang$^{1,2,3}$%
    \thanks{*Authors have equal contributions}
    \thanks{$^{1}$Interdisciplinary Program in AI, Seoul National University}%
    \thanks{$^{2}$Artificial Intelligence Institute, Seoul National University}%
    \thanks{$^{3}$Department of Computer Science, Seoul National University}%
    \thanks{This work was partly supported by Institute of Information \& communications Technology Planning \& Evaluation (IITP) grant funded by the Korea government(MSIT) [RS-2021-II211343, Artificial Intelligence Graduate School Program (Seoul National University)] and (RS-2021-II212068-AIHub/10\%, RS-2021-II211343-GSAI/15\%, 2022-0-00951-LBA/15\%, 2022-0-00953-PICA/20\%), NRF (RS-2024-00353991/20\%, RS-2023-00274280/10\%), and KEIT (RS-2024-00423940/10\%) grant funded by the Korean government.
    }%
}

\begin{document}
\maketitle
\thispagestyle{empty}
\pagestyle{empty}


\begin{abstract}
We introduce OV-MAP, a novel approach to open-world 3D mapping for mobile robots by integrating open-features into 3D maps to enhance object recognition capabilities. A significant challenge arises when overlapping features from adjacent voxels reduce instance-level precision, as features spill over voxel boundaries, blending neighboring regions together. Our method overcomes this by employing a class-agnostic segmentation model to project 2D masks into 3D space, combined with a supplemented depth image created by merging raw and synthetic depth from point clouds. This approach, along with a 3D mask voting mechanism, enables accurate zero-shot 3D instance segmentation without relying on 3D supervised segmentation models. We assess the effectiveness of our method through comprehensive experiments on public datasets such as ScanNet200 and Replica, demonstrating superior zero-shot performance, robustness, and adaptability across diverse environments. Additionally, we conducted real-world experiments to demonstrate our method's adaptability and robustness when applied to diverse real-world environments.
\end{abstract}


\section{INTRODUCTION}
In recent years, mobile robots have increasingly required robust 3D mapping capabilities to operate in complex, unstructured environments. Open-vocabulary 3D mapping, where the system can recognize and segment objects without being explicitly trained on specific object categories, plays a crucial role in enhancing these capabilities. The challenge of creating zero-shot 3D scene mappings from an open-world perspective arises from the complexity of 3D environments and the scarcity of large-scale open-world 3D datasets. Current approaches often rely on 2D open-vocabulary models, such as CLIP~\cite{radford2021learning}, to project features from RGB images into 3D space~\cite{huang2023visual, peng2023openscene, jatavallabhula2023conceptfusion, takmaz2023openmask3d}. By embedding CLIP-space features into the 3D space’s voxels, these methods aim to construct 3D maps that generalize to diverse environments.

However, many of these per-voxel mapping methods suffer from the problem where neighboring voxels share overly similar features, causing boundaries between instances to blur and reducing the precision of instance-level segmentation. This problem is especially pronounced when queried with a sentence or complex input, leading to poor object segmentation quality and unreliable mappings. Consequently, these approaches struggle to achieve the necessary instance-level precision required for accurate open-vocabulary 3D mapping.

\begin{figure} [t!]
\begin{center}
\includegraphics[width=\columnwidth]{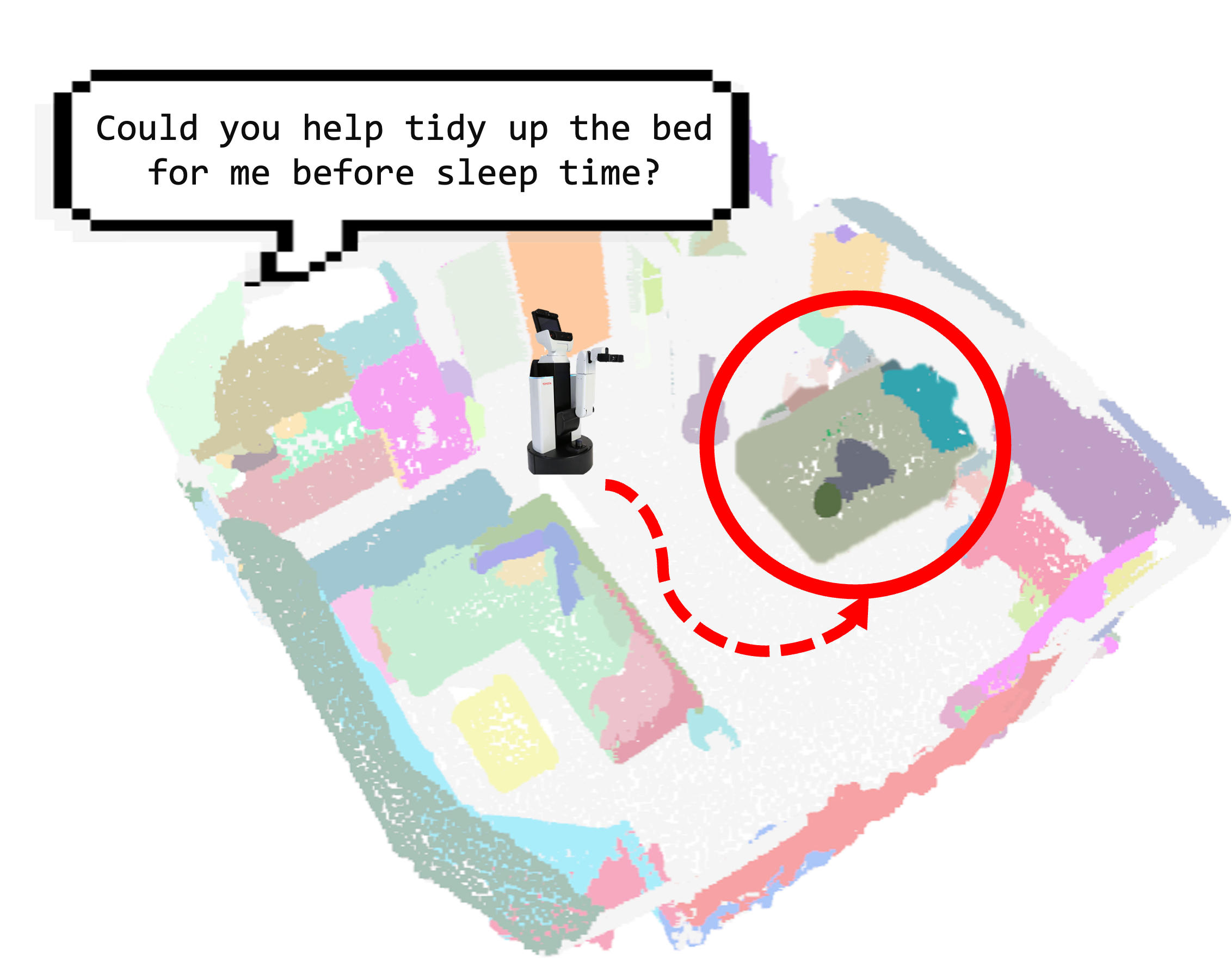}
\caption{\textbf{Illustration of Proposed System.} A mobile robot equipped with OV-MAP accurately identifies objects on a per-instance according to input queries.}
\label{system proposal}
\vspace{-7mm}
\end{center}
\end{figure}

To address these limitations, we propose OV-MAP (Open-vocabulary zero-shot 3D instance segmentation map), a novel zero-shot open-vocabulary 3D mapping method that operates from a per-instance perspective. This approach is inspired by advancements in zero-shot 3D segmentation~\cite{yang2023sam3d} and open-vocabulary 3D instance labeling~\cite{takmaz2023openmask3d}. OV-MAP confines CLIP features to individual 3D instances by developing a zero-shot 3D instance segmented map through a 3D mask voting process. The process begins by projecting 2D masks from class-agnostic segmentation models, such as those in~\cite{qi2022high, kirillov2023segment}, into 3D space using RGB-D images of the entire scene. These candidate 3D masks are then merged and refined through a voting mechanism applied over mesh-segmented areas, as described in~\cite{felzenszwalb2004efficient}. The dominant group of 3D masks is assigned to each area after the voting process.

\begin{figure*} [t!]
\begin{center}
\includegraphics[width=\textwidth]{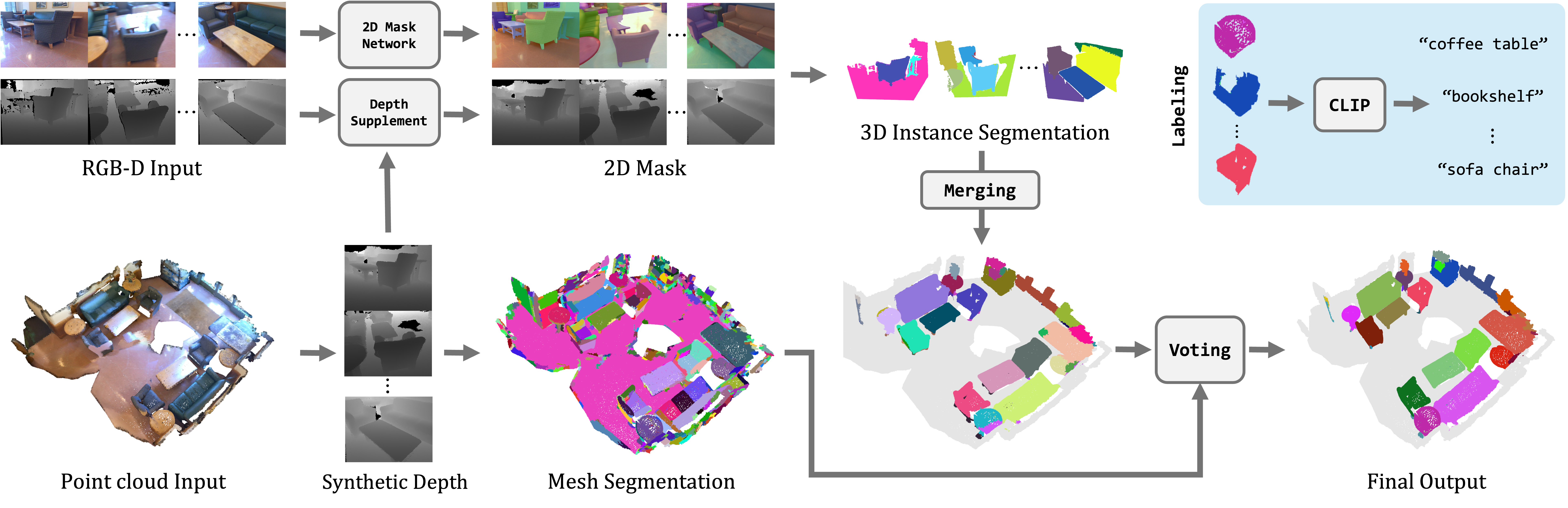} 
\caption{\textbf{Overview of OV-MAP Creation.} The pipeline begins with RGB-D input and point cloud data and progresses to the final 3D instance segmentation maps. Depth images are first refined with synthetic depth from point clouds. Next, RGB images are processed through a 2D class-agnostic segmentation network to generate 2D masks enriched with CLIP features. These masks are then projected into 3D space using supplemented depth data, forming preliminary 3D masks. The 3D masks are integrated with point cloud data to create 3D instance candidates, which are further refined in the final step using a voting mechanism and mesh segmentation to produce the final 3D instance segmentation maps.}
\label{system}
\end{center}
\end{figure*}

This process allows OV-MAP to generate 3D instance proposals without relying on 3D supervised learning models. Each instance is labeled using CLIP, based on the highest-scoring view from the 2D RGB images. A key advantage of this approach is its independence from supervised 3D instance segmentation models, making it adaptable to any scene equipped with RGB-D and point cloud data. Fig.~\ref{system proposal} illustrates an example of OV-MAP being used by a mobile robot to recognize objects based on input queries.

We verify OV-MAP through comprehensive experiments on publicly available datasets, including ScanNet200~\cite{rozenberszki2022language} and Replica~\cite{straub2019replica}, where our approach demonstrates robust zero-shot performance. Additionally, we apply OV-MAP in real-world environments, confirming its effectiveness beyond controlled dataset settings.

Our contributions are as follows:
\begin{itemize}
    \item We propose a mask voting method to merge 2D masks into 3D representations for open-vocabulary instance segmentation.
    \item We show that our per-instance zero-shot 3D mapping method enhances precision and generalization without relying on supervised 3D instance segmentation models.
    \item We validate OV-MAP on public datasets and real-world environments, demonstrating its versatility across various applications.
\end{itemize}

\section{Related Works}

In the domain of 3D open-vocabulary mapping, research can generally be categorized into two primary approaches: (1) leveraging 2D open-vocabulary models to create per-voxel 3D maps, and (2) employing trained 3D instance prediction models to generate per-instance 3D maps.

\subsection{Per-Voxel 3D Mapping}
In the context of 3D open-vocabulary mapping, per-voxel approaches rely on extracting features from 2D open-vocabulary models to generate 3D maps at the voxel level. Recent advancements in 2D open-vocabulary segmentation models~\cite{li2022language, liang2023open, qin2023freeseg, ghiasi2022scaling} have laid the groundwork for exploring 3D open-vocabulary segmentation~\cite{huang2023visual, peng2023openscene, gadre2022clip, jatavallabhula2023conceptfusion, ding2023pla, ha2022semantic}. Projects such as OpenScene~\cite{peng2023openscene} and ConceptFusion~\cite{jatavallabhula2023conceptfusion} leverage these 2D open-vocabulary models \cite{li2022language, liang2023open} to extract per-pixel features, which are then projected onto 3D scene voxels to build open-vocabulary representations. Despite these advances, representing 3D scenes at the voxel level presents challenges, particularly the issue where features overflow between voxel boundaries, reducing the accuracy of open-vocabulary queries and making it difficult to distinguish between adjacent instances.

\subsection{Per-Instance 3D Mapping}
To overcome the limitations of per-voxel methods, per-instance approaches focus on generating 3D maps at the instance level. OpenMask3D~\cite{takmaz2023openmask3d}, for instance, proposes a method that generates 3D instance proposals using Mask3D~\cite{schult2023mask3d} and enriches each instance with CLIP features extracted from 2D RGB images. This approach has shown potential for improving mapping accuracy in datasets like ScanNet~\cite{dai2017scannet}. However, one limitation of OpenMask3D is that it relies on Mask3D, a 3D supervised model, which struggles to generalize to novel scenes, leading to suboptimal 3D instance predictions. Our work addresses this issue by introducing a new per-instance mapping approach that enhances generalization across diverse environments. Specifically, we integrate outputs from 2D class-agnostic segmentation model~\cite{qi2022high} into the 3D space, similar to the method used in SAM3D~\cite{yang2023sam3d}.

\section{Method} \label{method}
 
The process we've developed is outlined in Fig.~\ref{system}. Starting with RGB-D images from a scene and its reconstructed point cloud, our first step involves generating a list of 3D candidate masks. Each mask is uniquely identified by group IDs generated using class-agnostic 2D pre-trained segmentation models (Sec.~\ref{method}-A). Following this, we initiate a merging process, where similar 3D masks bearing different group IDs are merged under unified group IDs. After this merging phase, we engage in a dominant voting process, allocating the leading 3D mask groups to respective mesh segmented areas (Sec.~\ref{method}-B). This results in final group IDs that are directly associated with individual 3D instances within the scene. In the final step, we enrich each 3D instance with the top-scoring per-mask open-vocabulary CLIP features derived from 2D RGB images (Sec.~\ref{method}-C). This comprehensive approach ensures each 3D instance is accurately defined and enriched within the scene.

\subsection{Candidate 3D Masks Generation}
 
The first step in our approach involves extracting accurate 2D masks for each RGB image using a 2D class-agnostic segmentation model. These masks are then projected onto the 3D surface points of the scene's point cloud, generating 3D masks with unique group IDs.

\textbf{2D Mask Generation.} Given RGB images with a resolution of \( H \times W \), denoted as \({I}^{RGB}_{t}\), we obtain 2D masks from the 2D segmentation model \({S}^{2D}({I}^{RGB}_{t})\), denoted as \({m}^{2D}_{t} \in \mathbb{R}^{H \times W \times M}\), where \(M\) is the predicted masks, \(t\) is the sequence of total frames in the scene. For \({S}^{2D}\), we experiment with the class-agnostic segmentation model, CropFormer~\cite{qi2022high}.

\begin{figure} [t!]
\begin{center}
\includegraphics[width=\columnwidth]{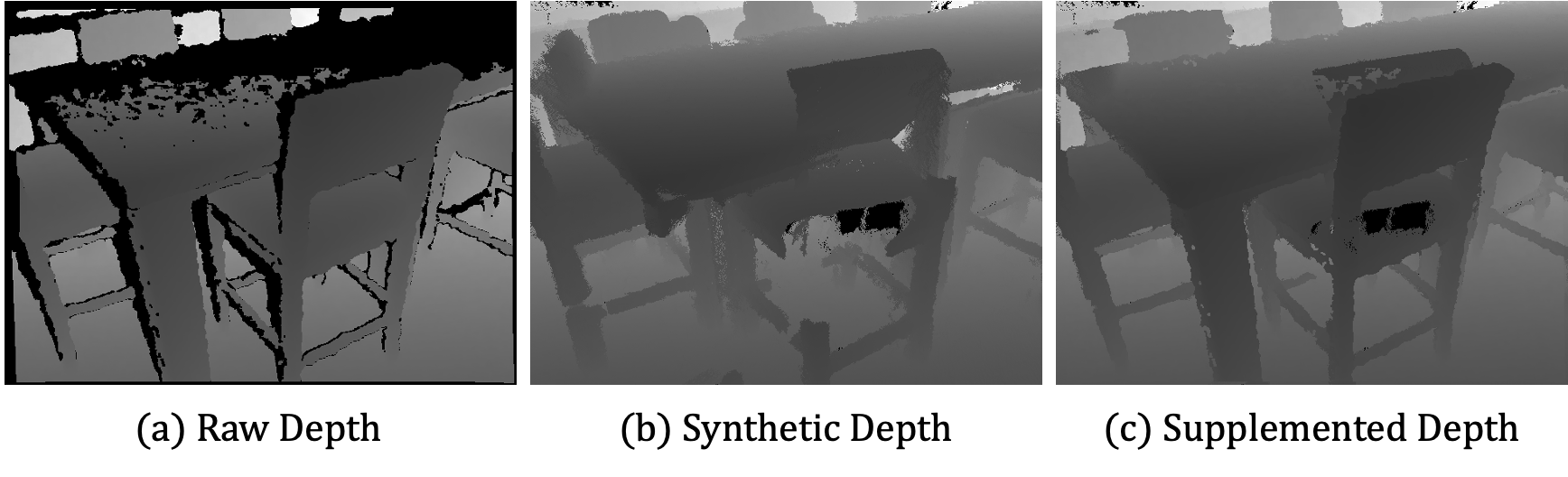}
\caption{\textbf{Supplemented Depth Image Construction.} The original depth image (a) is enhanced using the synthetic image (b), producing the supplemented depth image (c).}
\vspace{-6mm}
\label{depth_merge}
\end{center}
\end{figure}

\textbf{Depth Image Construction.} To project ${m}^{2D}_{t}$ to the 3D point cloud space, depth images are required. However, real-world depth images do not guarantee perfect depth information for every pixel, such as on glass surfaces. To address this, we supplement the missing parts of the depth image, ${I}^{d}_{t}$, with information from a synthetic depth image, ${I}^{d'}_{t}$, derived from a fully constructed point cloud ${P}$. The supplementation process can be described as follows:

Given a raw depth image ${I}^{d}_{t}$ and a synthetic depth image ${I}^{d'}_{t}$, the supplemented depth image ${I}^{sd}_{t}$ is constructed by:
\begin{equation}
{I}^{sd}_{t}(i, j) = 
\begin{cases} 
{I}^{d'}_{t}(i, j) & \text{if } {I}^{d}_{t}(i, j) = 0 \text{ or } {I}^{d'}_{t}(i, j) = 0 \\
{I}^{d}_{t}(i, j) & \text{otherwise}
\end{cases}
\end{equation}
for each pixel coordinate $(i, j)$. This approach ensures that the depth image incorporates reliable depth information from the synthetic image where the original depth image is lacking, as shown in Fig.~\ref{depth_merge}.

\textbf{3D Mask Generation.} Given ${I}^{sd}_{t}$ and ${P}$, and a specific frame index $t$, we can generate multiple 3D masks, $m^{3D}_{t,g} \in \mathbb{R}^{L \times 3}$, each with $L$ points and a unique group identifier $g$. This is achieved by projecting a 2D mask, $m^{2D}_{t}$, onto ${P}$ using the corresponding depth images, ${I}^{sd}_{t}$. To facilitate this projection, we first establish a grid of pixel coordinates, $\mathbf{u} = (u, v)$, that matches the resolution of ${I}^{sd}_{t}$. These coordinates are then merged with the depth information to create a set of 3D points in the camera coordinate system, represented as $(u, v, \frac{I^{sd}(u, v)}{s})$, where $s$ is the scaling factor for depth. Subsequently, the inverse of the intrinsic camera matrix, $K$, is applied to these points to transition them to camera space coordinates. The final step involves transforming these camera space points to world coordinates by applying the pose transformation matrix, $T$, thereby yielding the 3D mask, $m^{3D}_{t,g}$, that correlates with the original 2D mask, $m^{2D}_{t}$.

\subsection{Dominant Group Voting for 3D Instance Segmentation}
 
Upon obtaining candidate 3D masks, denoted by $m^{3D}_{t,g}$, projected onto the point cloud ${P}$, we embark on a process of merging these masks and voting to ascertain the dominant area within each mesh segmentation based on~\cite{felzenszwalb2004efficient}. This approach facilitates the identification of dominant groups within the segmentation meshes. As a result, each mesh segmentation is attributed a unique dominant group identifier, $g$, distinguishing it as a separate entity within the 3D instance segmentations.

\textbf{Merging.} To delve deeper into the merging process, consider $m^{3D}_{t,g}$, where we combine every two frames of a point cloud from $2t$ and $2t+1$ frames to create $t'$ combinations, effectively halving the length of $t$. Within this setup, each pair of unique group identifiers, $g'$ and $g''$, in $m^{3D}_{t',g}$, is evaluated for its overlapping rate $OR$. This rate determines whether the groups $g'$ and $g''$ should be merged into the same group $g'$. The $OR$ is calculated as follows:
\begin{equation}
    OR = \frac{\text{overlapping}(m^{3D}_{t', g'}, m^{3D}_{t', g''})}{\max(\text{len}(m^{3D}_{t', g'}), \text{len}(m^{3D}_{t', g''}))}
\end{equation}
If the $OR$ for a smaller group surpasses a predefined threshold, then group $g''$ is merged into group $g'$, represented by the following:
\begin{equation}
    \text{if } OR > \text{threshold} \text{, then assign } m^{3D}_{t', g''} \text{ to } m^{3D}_{t', g'}.
\end{equation}

\begin{table*}[t!]
\centering
\caption{3D instance segmentation Results on ScanNet200~\cite{rozenberszki2022language}.}
\label{tab:3d-instance-segmentation}
\begin{tabular}{@{}lccccccccc@{}} 
\toprule
\textbf{Model} &\textbf{Open-Vocab} & \textbf{3D Proposal} & \textbf{Map Type} & \textbf{AP} & \textbf{AP$_{50}$} & \textbf{AP$_{25}$} & \textbf{head (AP)} & \textbf{common (AP)} & \textbf{tail (AP)} \\ \midrule
Mask3D~\cite{schult2023mask3d}& & Supervised & Per-Instance & 26.9 & 36.2 & 41.4 & 39.8 & 21.7 & 17.9 \\
OpenMask3D~\cite{takmaz2023openmask3d} & \checkmark & Mask3D~\cite{schult2023mask3d} & Per-Instance & 15.4 & 19.9 & 23.1 & 17.1 & 14.1 & 14.9 \\
\midrule
OpenScene~\cite{peng2023openscene} & \checkmark & - & Per-Voxel & 6.6 & 10.2 & 14.8 & 7.2 & 5.8 & 6.9 \\
SAM3D~\cite{yang2023sam3d}& \checkmark & None & Per-Instance & 8.4 & 13.1 & 18.7 & 9.3 & 7.0 & 9.1 \\
Ours & \checkmark & None & Per-Instance & \textbf{11.9} & \textbf{17.4} & \textbf{23.2} & \textbf{12.5} & \textbf{10.5} & \textbf{12.7} \\
\bottomrule
\end{tabular}
\end{table*}

\begin{table}[t!]
\centering
\caption{3D instance segmentation results on Replica dataset~\cite{straub2019replica}.}
\label{replica_exp}
\begin{tabular}{@{}lccc@{}}
\toprule
\textbf{Model} & \textbf{AP} & \textbf{AP$_{50}$} & \textbf{AP$_{25}$}  \\ \midrule
Mask3D~\cite{schult2023mask3d}       & 5.8   & 8.5       & 10.7        \\
OpenMask3D~\cite{takmaz2023openmask3d}   & 13.1   & 18.4       & 24.2       \\
OpenScene~\cite{peng2023openscene}      & 7.3   & 9.4      & 11.2       \\
Ours         & \textbf{14.2}   & \textbf{19.6 }     & \textbf{28.1}       \\
\bottomrule
\end{tabular}
\end{table}

Our merging criterion is specifically designed to differ from methods used by others, like~\cite{yang2023sam3d}, which merges groups based predominantly on the overlapping rate of smaller groups. Although this approach can result in a higher rate of group consolidation, it bears the drawback of larger 3D masks potentially subsuming smaller ones, consequently eroding the granularity of detail. Moreover, inaccuracies in mask generation, particularly those arising from blurred RGB images, can critically degrade the quality of segmentation when larger masks unduly influence the merging process.

Given these factors, our method prioritizes calculating the $OR$ by concentrating on larger masks to ensure that only those with substantial similarity are combined. This careful approach preserves the detail within smaller masks and lessens the chances of errors when merging into larger masks. Additionally, we refine the resolution of the 3D mask area by voxelizing merged points at a quarter of the original voxel scale. This step is key to preserving clear group dominance within the 3D space, which in turn, enhances the dominant group voting process—a critical component for the next stage of segmentation.

\textbf{Dominant Voting.} After merging $t$ frames into a comprehensive 3D mask \(m^{3D}_{g} \in {P}\), with each mask having unique identifiers $g$ for the entire scene's 3D masks, we identify mesh segmented areas, denoted as \(ms_{a} \in {P}\), where \(a\) represents each distinct segmented area. These areas are determined through~\cite{felzenszwalb2004efficient}, with the calculation of \(ms_{a}\) based on surface normals. This method positions \(ms_{a}\) as the foundational units for our 3D instance segmentations, marking them as the primary representatives. Subsequently, each area in \(ms_a\) is assigned the most dominant group id \(g\), determined through a voting process within the area, formalized by the following equation:
\begin{equation}
    ms_{a'} = \text{vote}(ms_{a} \cap m^{3D}_{g})
\end{equation}

Finally, with \(ms_{a}\) assigned the dominant group id \(g\), it becomes the definitive outcome of our 3D instance segmentation effort, with each \(ms_{a'}\) representing an individual instance.

\subsection{Per-Mask Open-Vocabulary Embedding}
During the process of obtaining the final 3D instance segmentation \(ms_{a}\), we assign 3D mask scores to each group \(g\) within \(m^{3D}_{t,g}\). These scores are calculated based on the pixel counts in the RGB image \({I}^{RGB}_{t}\) and the point count in the point cloud \({P}\), as defined by:
\begin{equation}
    \text{Score}_{t,g} = \alpha \cdot \text{PixelCount}({I}^{RGB}_{t}, g) + \beta \cdot \text{PointCount}({P}, g)
\end{equation}
where \(\alpha\) and \(\beta\) are weighting factors that modulate the contributions of pixel and point counts to the total score of each 3D mask.

In the merging process, if two group ids are combined, the higher score, \(\text{Score}_{t,g}\), is preserved for the unified group \(g\). In the finalization stage of the 3D instance segmentation, the highest \(\text{Score}_{t,g}\) within each group is leveraged to select targeted frames from \({I}^{RGB}_{t}\). This selection process enables the extraction of 2D image crops corresponding to the \(g\) mask area. Subsequently, these crops are embedded with CLIP features for open-vocabulary inference, enhancing the semantic richness and applicability of the segmentation in diverse analytical contexts.

\section{Experiments} \label{experiments}

This section outlines our comprehensive assessment of the proposed 3D instance segmentation approach. We begin by detailing the datasets and evaluation metrics employed in our analysis (Sec.~\ref{experiments}-A). Subsequent sections delve into statistical analysis of segmentation performance (Sec.~\ref{experiments}-B), presentation of qualitative outcomes (Sec.~\ref{experiments}-C), a comparative ablation study analyzing the effects of different depth data types on segmentation performance (Sec.~\ref{experiments}-D), and real-world experiments to demonstrate the practical applicability of our method (Sec.~\ref{experiments}-E).

\subsection{Experimental Setting}

\textbf{Datasets.} Our evaluation leverages the ScanNet200 \cite{rozenberszki2022language} and Replica~\cite{straub2019replica} datasets, offering a diverse set of environments for assessing our method's efficacy. Specifically, the validation subset of ScanNet200, comprising 312 scenes annotated across 200 categories, serves as the primary benchmark for our 3D instance segmentation task. The categories are divided into head (66 categories), common (68 categories), and tail (66 categories) subsets based on label frequency, facilitating an in-depth analysis across varying data distributions. Furthermore, the inclusion of the Replica dataset, with its detailed office and room scenes, aids in evaluating our method's zero-shot ability across different settings.

\begin{figure} [t!]
\begin{center}
\includegraphics[width=\columnwidth]{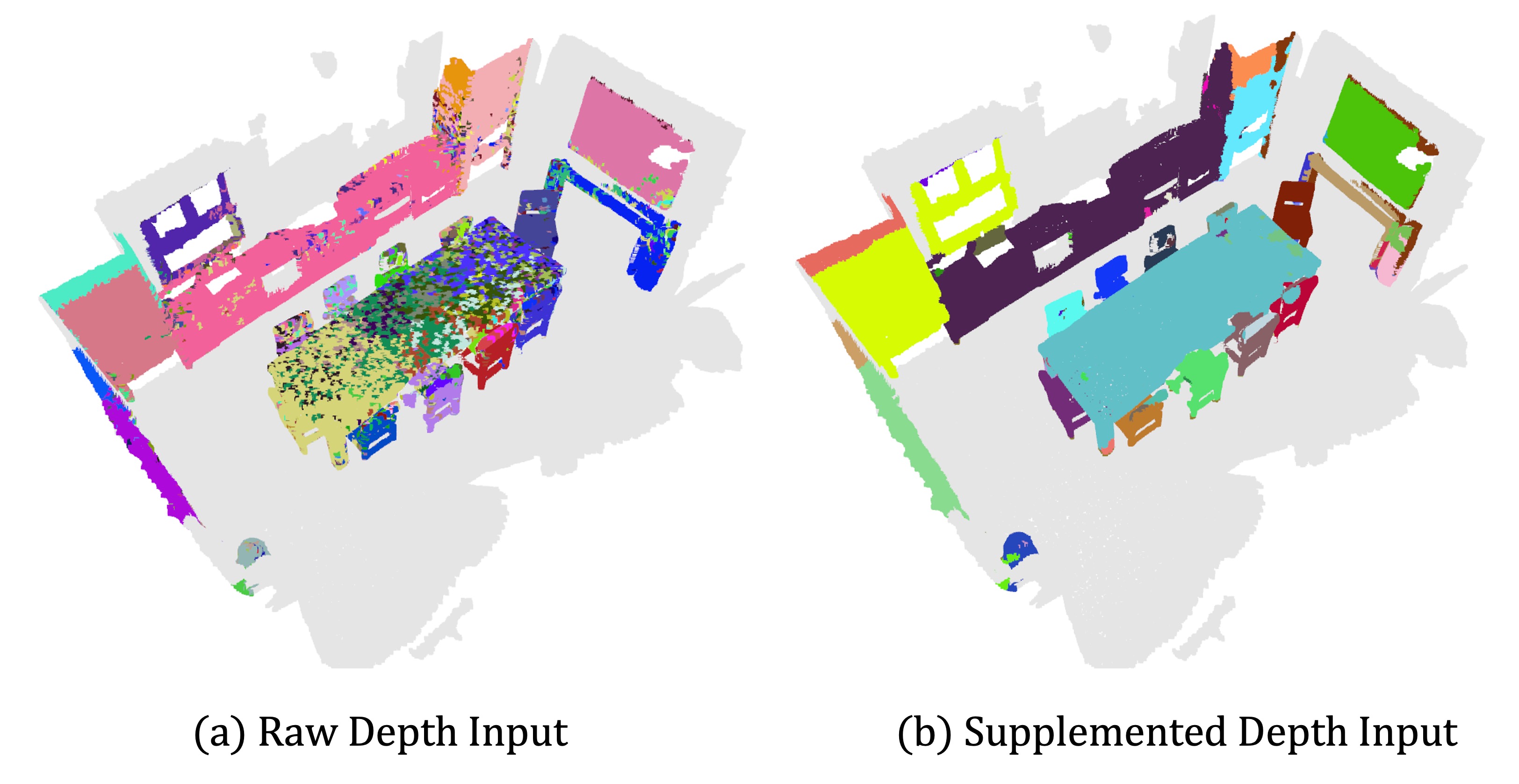}
\caption{\textbf{Impact of Using Supplemented Depth Data.} Comparison of results from raw depth data (a) against supplemented depth data (b) for 3D map construction, specifically examining outputs produced prior to the implementation of the dominant voting process.}
\label{depth_merge_exp}
\vspace{-7mm}
\end{center}
\end{figure}

\begin{figure*} [t!]
\begin{center}
\includegraphics[width=\textwidth]{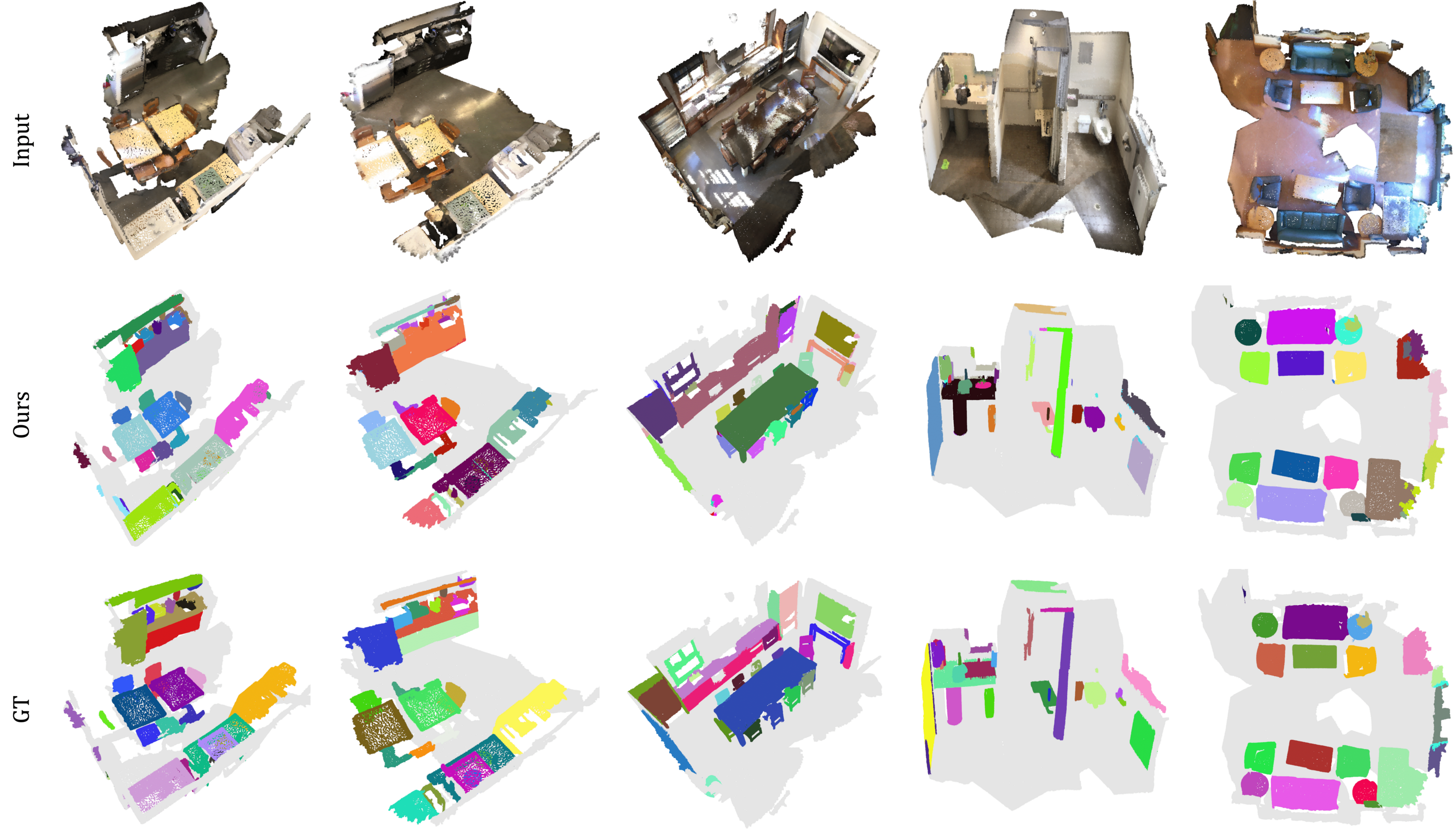} 
\caption{\textbf{Qualitative Analysis of Segmentation Performance.} The images show input scenes, Ground Truth (GT) annotations for each scene, and our model's results demonstrating the effectiveness of open-vocabulary instance segmentation.}
\label{q_result}
\vspace{-5mm}
\end{center}
\end{figure*}

\textbf{Metrics.} To quantify the segmentation accuracy, we adopt the average precision (AP) metric, a standard in 3D instance segmentation evaluations. AP calculations are conducted at two mask overlap thresholds—50\% and 25\%—and results are averaged over a range from 0.5 to 0.95 in increments of 0.05, as per ScanNet's evaluation protocol~\cite{rozenberszki2022language}. Each predicted mask in our model is ascribed a uniform prediction confidence score of 1.0 for simplicity in metric computation.

\textbf{Implementation details.} Our system processes posed RGB-D pairs from ScanNet200 and Replica datasets, analyzing 1 frame out of every 10 in RGB-D sequences. To obtain object-level masks, we employ CropFormer~\cite{qi2022high} as our 2D mask predictor. For open-vocabulary feature extraction, we use CLIP ViT-H. We voxelize the reconstructed point cloud with a radius of 2cm and merge voxelization with a radius of 0.05cm for ScanNet and Replica. We perform post-processing to refine the output 3D mask by using the nearest neighbor algorithm~\cite{taunk2019brief} and DBSCAN~\cite{schubert2017dbscan} to separate disconnected point clusters and remove irrelevant ones.

\subsection{Quantitative Results}

\textbf{Evaluation on ScanNet200.} We present our evaluation results for the ScanNet200 dataset, detailing AP, AP50, and AP25 metrics for the dataset's categories in Tab.~\ref{tab:3d-instance-segmentation}. Our method outperforms traditional per-voxel mapping approaches, highlighting the efficacy of our zero-shot 3D instance segmentation strategy. Notably, methods such as Mask3D~\cite{schult2023mask3d} and OpenMask3D~\cite{takmaz2023openmask3d} exhibit better performance, benefiting from direct learning on the ScanNet dataset.

While supervised methods generally excel in the `head' and `common' categories, leveraging the extensive training on these more frequent categories, they tend to falter in the `tail' category, where instances are less common. This contrast underscores a limitation in the adaptability of supervised approaches to the long-tail distribution of dataset categories. In comparison, our method maintains consistent performance across all categories—head, common, and tail—demonstrating its robustness and general applicability for 3D instance segmentation across a diverse range of scenes.

\textbf{Evaluation on Replica.} We performed a zero-shot evaluation of our method on the Replica dataset to assess its generalization potential. The results, detailed in Tab.~\ref{replica_exp}, show our method significantly outperforming the baseline models. This performance disparity highlights the limitations of models reliant on ScanNet-specific training when faced with the Replica dataset, a different environment. Such a decline in performance underscores the importance of further developing zero-shot 3D instance mapping techniques. Our findings advocate for methods with higher adaptability and underscore the need for models that can seamlessly generalize across a variety of datasets, ensuring wide-ranging applicability and a deeper understanding of diverse environments.

\begin{figure*} [t!]
\begin{center}
\includegraphics[width=0.83\textwidth]{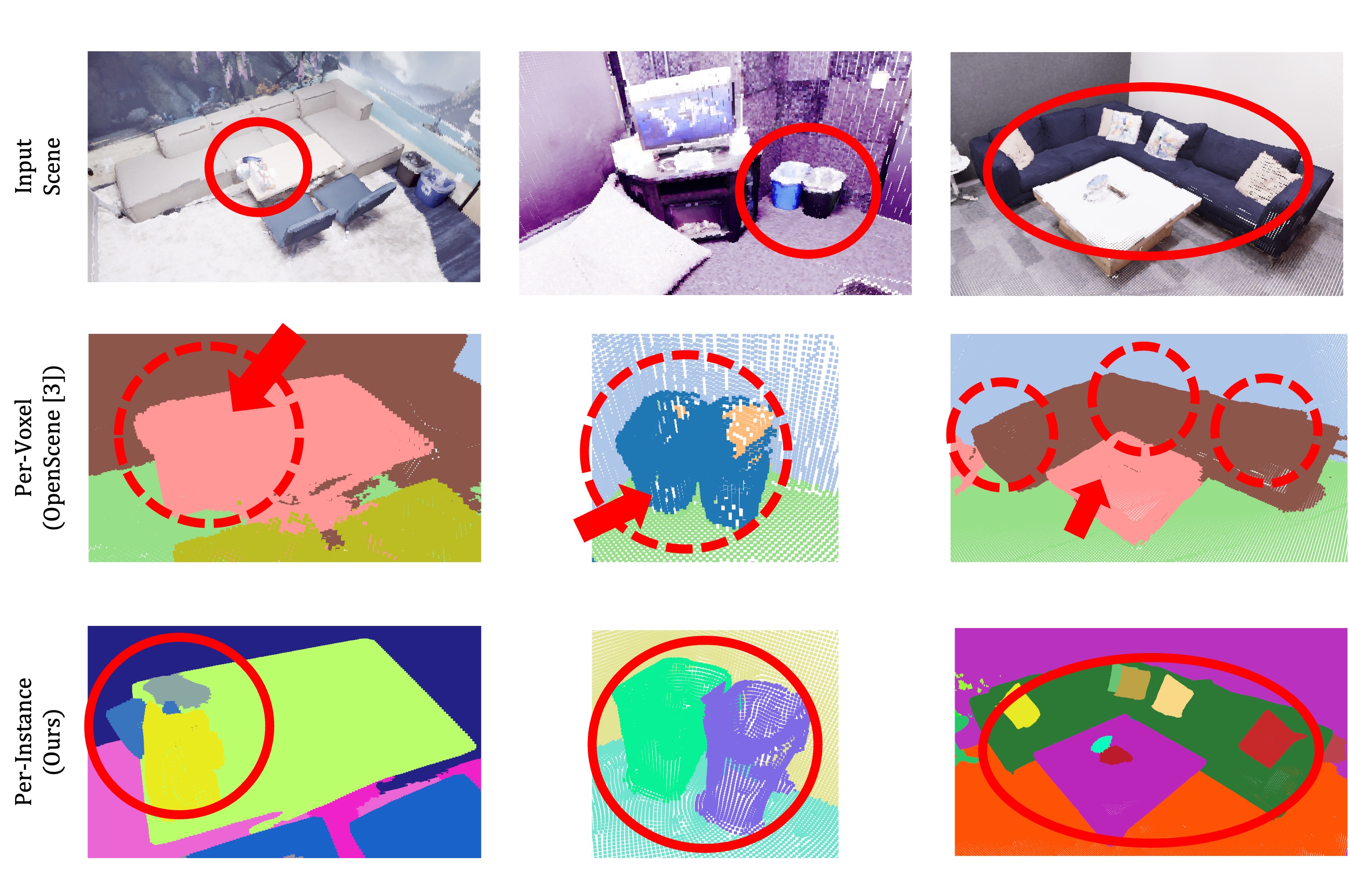} 
\caption{\textbf{Qualitative Results of .} Per-Voxel method~\cite{peng2023openscene} vs. Per-Instance method (Ours). Unlike the per-voxel method, our approach segments nearby objects, such as items on a desk, an attached trash bin, and cushions on a sofa.
The white dashed
circles indicate the most noticeable differences between both approaches. Color and ground truth are shown
for reference only. Overall, our approach produces less noisy segmentation masks.
}

\label{c_result}
\end{center}
\end{figure*}

\subsection{Qualitative Results}

As shown in Fig.~\ref{depth_merge_exp}, incorporating additional depth information from point clouds is crucial for the transition from 2D to 3D segmentation techniques. While precise 2D segmentation is foundational, its effectiveness may be diminished when depth images are affected by reflective objects, potentially hindering the 2D to 3D conversion process. Therefore, enriching our depth data with scenes reconstructed from point clouds plays a vital role in achieving reliable 2D to 3D mapping.

\begin{table}[t!]
\centering
\caption{Effect of Different Depth Image Types on 3D Instance Segmentation~\cite{rozenberszki2022language}.}
\label{depth_ablation_exp}
\begin{tabular}{@{}lccc@{}}
\toprule
\textbf{Depth Type} & \textbf{AP} & \textbf{AP$_{50}$} & \textbf{AP$_{25}$}  \\ \midrule
Raw depth       & 11.1   & 17.1       & 22.5        \\
Synthetic depth   & 10.1   & 15.9       & 20.7       \\
Supplemented depth  & \textbf{11.9}   & \textbf{17.4}     & \textbf{23.2}       \\ \bottomrule
\end{tabular}
\end{table}

Further, in Fig.~\ref{q_result}, we present qualitative examples of our approach applied to the task of open-vocabulary 3D instance segmentation. Leveraging its zero-shot capabilities, OV-MAP generates high-quality 3D instance segmentation maps where every instance is amenable to open-vocabulary queries. Remarkably, OV-MAP achieves this level of accuracy without relying on models trained specifically on the ScanNet200 dataset, showcasing its effectiveness and the potential for wide applicability in diverse settings.

Furthermore, Fig.~\ref{c_result} showcases the distinguishable advantages of our method over the traditional per-voxel mapping approach~\cite{peng2023openscene}. The per-voxel method often conflates adjacent instances, leading to a less accurate representation of the scene. In contrast, our utilization of a 2D class-agnostic model for segmentation demonstrates superior performance in distinguishing between individual instances. This approach not only enhances the clarity of the mapping results but also underscores the effectiveness of class-agnostic models in achieving precise instance segmentation within 3D spaces.

\begin{figure} [t!]
\begin{center}
\includegraphics[width=\columnwidth]{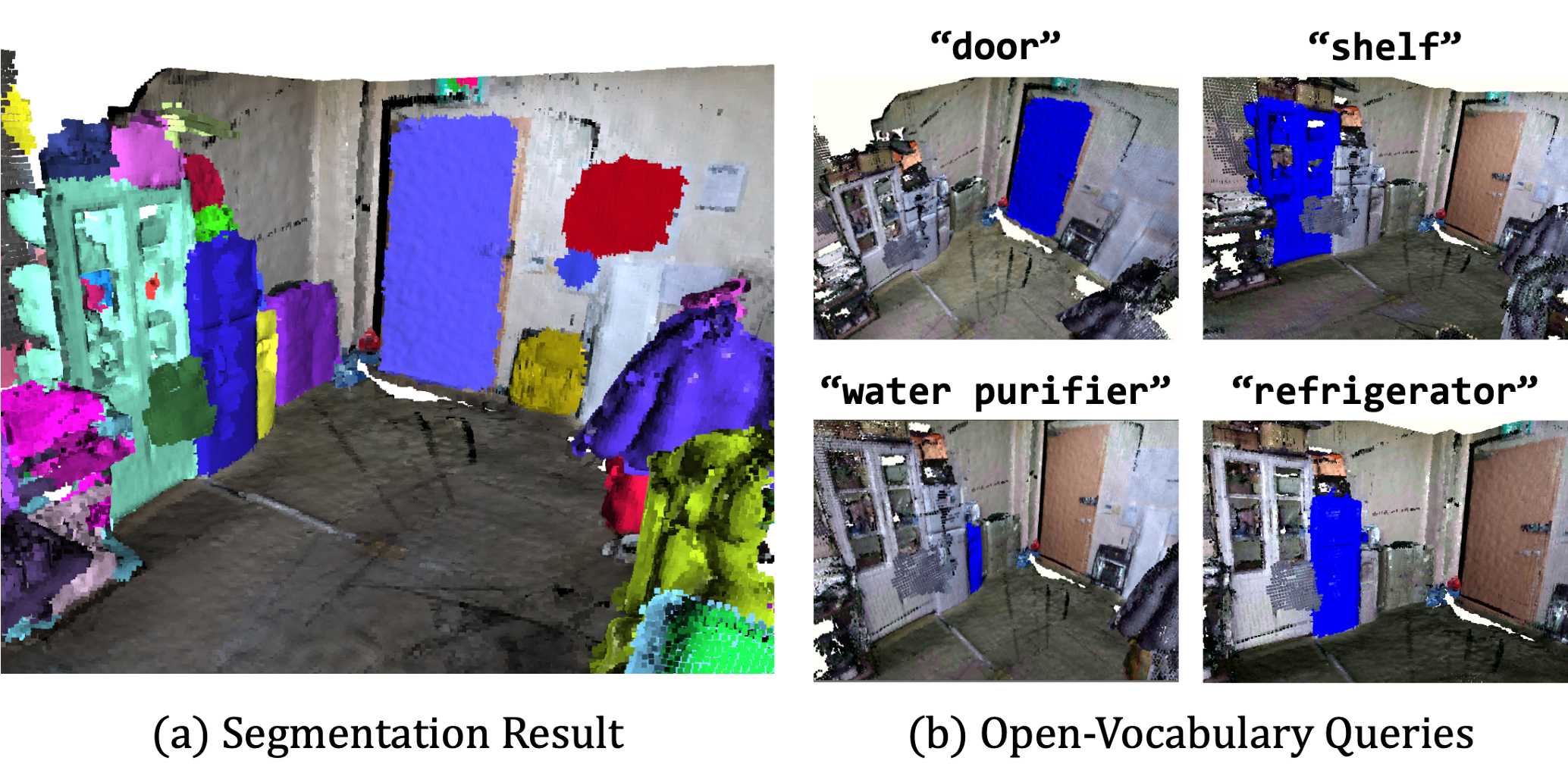}
\caption{\textbf{Real-World Map Creation Validation.} (a) 3D instance segmentation on real-world data. (b) Open-vocabulary query results showing segmented objects.}
\label{real_ex}
\vspace{-3mm}
\end{center}
\end{figure}

\subsection{Ablation Study}
To evaluate the impact of different depth data types on 3D mask generation, we conducted an ablation study comparing raw, synthetic, and supplemented depth images. As shown in Table~\ref{depth_ablation_exp}, the use of supplemented depth data, which combines both raw and synthetic sources, significantly improves performance across all metrics. For instance, the AP increases from 11.1 (raw) to 11.9 (supplemented), with corresponding gains in AP${50}$ and AP${25}$. This highlights the effectiveness of using enhanced depth data to improve 2D-to-3D mask projections, resulting in more accurate 3D instance segmentation.

\subsection{Real-World Experiments}

To validate the real-world applicability of our approach, we captured RGB-D data from an actual indoor environment and processed it to generate detailed point clouds of the scene. Our method was then applied to accurately identify and segment individual objects based on open-vocabulary queries. The results of the instance segmentation are shown in Fig.~\ref{real_ex}(a), while the corresponding open-vocabulary query results are illustrated in Fig.~\ref{real_ex}(b). Notably, objects such as `water purifier,' `door,' `shelf,' and `refrigerator' were correctly segmented and matched purely from open-vocabulary descriptions, without relying on predefined labels. These results highlight that our OV-MAP method not only achieves accurate instance segmentation but also excels in handling open-vocabulary inputs, demonstrating its effectiveness and potential for use in diverse, real-world environments.

\section{Conclusion}
In this paper, we presented OV-MAP, an open-vocabulary zero-shot 3D instance segmentation map for mobile robots. By employing a class-agnostic segmentation model for accurate 2D-to-3D mask projection and incorporating a 3D mask voting mechanism, our method achieves zero-shot 3D instance mapping. It demonstrates superior performance through rigorous testing on the ScanNet200 and Replica datasets. This work represents a significant advancement in open-vocabulary instance segmentation for 3D scenes and has potential applications in mobile robotics.

\bibliographystyle{ieeetr}
\bibliography{ref}
\end{document}